\documentclass[10pt,twocolumn,letterpaper]{article}

\usepackage{iccv}
\usepackage{times}
\usepackage{epsfig}
\usepackage{graphicx}
\usepackage{amsmath}
\usepackage{amssymb}
\usepackage{comment}
\usepackage{footnote}
\usepackage{epstopdf}
 \usepackage{enumitem}
\usepackage{subfigure}
\usepackage{caption}
\usepackage{comment}
\usepackage{multirow}
\usepackage{algorithm}
\usepackage{times}
\usepackage{epsfig}
\usepackage{graphicx}
\usepackage{amsmath}
\usepackage{amssymb}
\usepackage[noend]{algpseudocode}
\usepackage{wrapfig}
\usepackage{amsthm}  
\usepackage{wasysym}
\usepackage[ruled,vlined,algo2e]{algorithm2e}

\providecommand{\SetAlgoLined}{\SetLine}
\usepackage{color}
\usepackage{dsfont}	
\newcommand{\bfsection}[1]{\vspace*{0.1cm}\noindent\textbf{#1.}}

\usepackage[breaklinks=true,bookmarks=false]{hyperref}

\iccvfinalcopy 

\def\iccvPaperID{****} 

\ificcvfinal\fi

\begin{document}

\title{ A Technical Survey and Evaluation of Traditional Point Cloud Clustering Methods for LiDAR Panoptic Segmentation}

\author{Yiming Zhao \hspace{2cm} Xiao Zhang \hspace{2cm} Xinming Huang\\
Worcester Polytechnic Institute\\

100 Institute Rd, Worcester, MA, USA\\

{\tt\small \{yzhao7, xzhang25, xhuang\}@wpi.edu}


}

\maketitle
\ificcvfinal\thispagestyle{empty}\fi

\begin{abstract}
   LiDAR panoptic segmentation is a newly proposed technical task for autonomous driving. In contrast to popular end-to-end deep learning solutions, we propose a hybrid method with an existing semantic segmentation network to extract semantic information and a traditional LiDAR point cloud cluster algorithm to split each instance object. We argue geometry-based traditional clustering algorithms are worth being considered by showing a state-of-the-art performance among all published end-to-end deep learning solutions on the panoptic segmentation leaderboard of the SemanticKITTI dataset. To our best knowledge, we are the first to attempt the point cloud panoptic segmentation with clustering algorithms. Therefore, instead of working on new models, we give a comprehensive technical survey in this paper by implementing four typical cluster methods and report their performances on the benchmark. Those four cluster methods are the most representative ones with real-time running speed. They are implemented with C++ in this paper and then wrapped as a python function for seamless integration with the existing deep learning frameworks. We release our code for peer researchers who might be interested in this problem \footnote{https://github.com/placeforyiming/ICCVW21-LiDAR-Panoptic-Segmentation-TradiCV-Survey-of-Point-Cloud-Cluster}. 
     
\end{abstract}

\section{Introduction}

Panoptic segmentation is an ensemble of both the semantic segmentation for static stuff and the instance segmentation for countable objects. Point cloud panoptic segmentation needs to provide the semantic label id for each point and further assign a unique instance label to points that belong to the same object \cite{milioto2020lidar}. With the understanding of both the semantic and instance information from a single frame LiDAR scan, this task is able to deliver many useful clues for advanced autonomous driving functions, such as future prediction \cite{hendy2020fishing} and map building \cite{chen2019suma++}.      

\begin{figure}[t]
    \includegraphics[width=\linewidth, height=5.5cm]{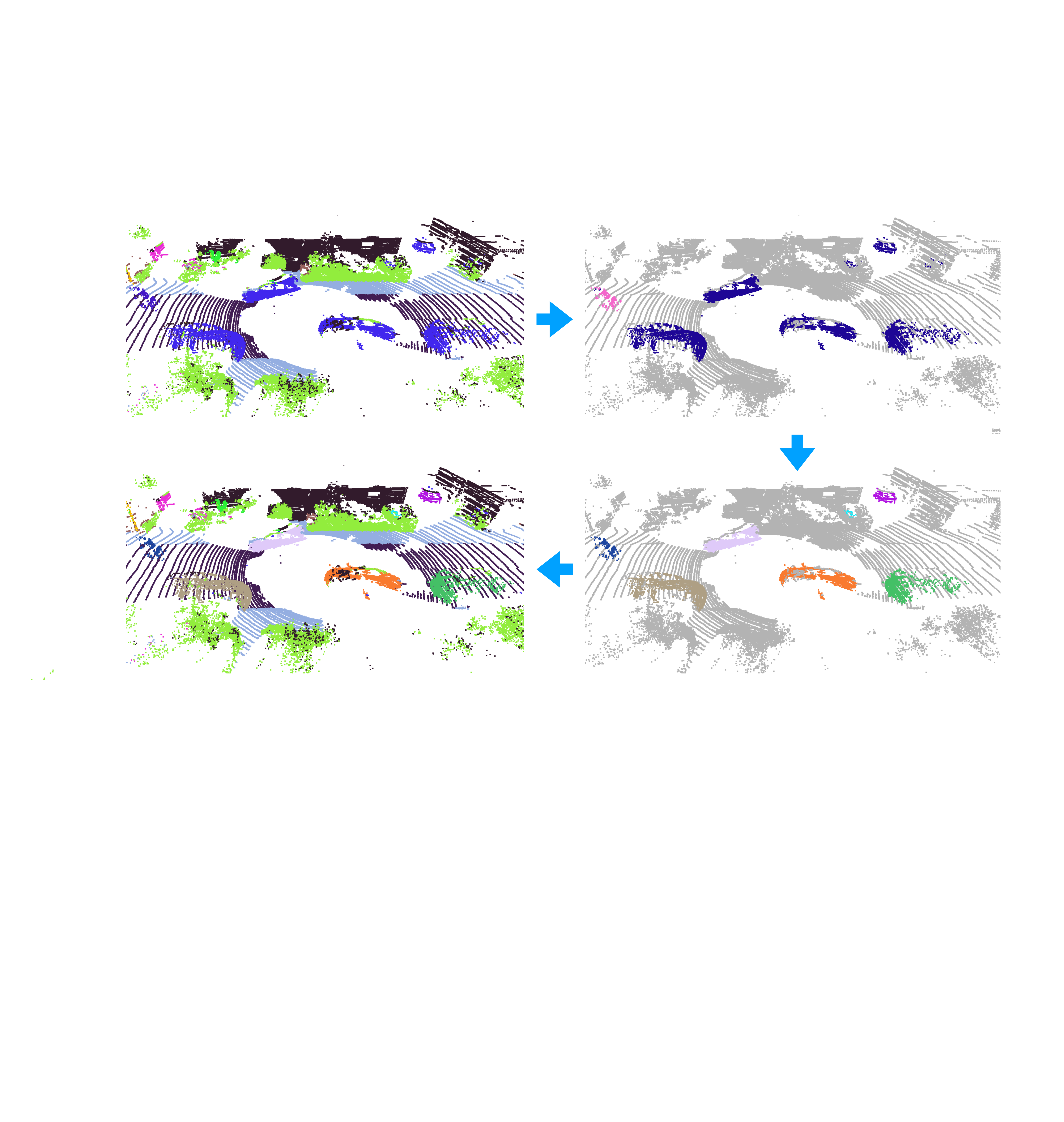}
    \caption{The demonstration of the panoptic segmentation task and how our pipeline solves it. After the semantic segmentation, the clustering algorithm works on object points to further segment each instance.}
    \vspace{-0.4cm}
    \label{fig:SLR_Cluster_example}
\end{figure}

\bfsection{Motivation} Most existing works solve the LiDAR panoptic segmentation with end-to-end deep learning models. A neural network is in charge of both the point-wise classification and the point-wise clustering. Nowadays, almost all classification tasks are dominated by neural networks. However, considering the 3D geometry information implicitly encoded in the point cloud, it is questionable if the neural network is good at clustering at the same time. At least, instead of directly applying the neural network as the solution, we should evaluate several questions. For instance, do we really need the neural network for point-wise clustering? If needed, how much the neural network solution can outperform traditional methods? Answer those questions need a solid benchmark to investigate and compare existing traditional clustering methods on well-recognized datasets. This motivates us to prepare this technical survey paper of the traditional point cloud clustering method on the SemanticKITTI point cloud panoptic segmentation task.

Point cloud clustering is a topic studied in multiple domains, including robotics \cite{zermas2017fast} and intelligent transportation \cite{shin2017real}. This technique has been used in various applications. For example, some recent point cloud compression methods need clustering algorithms to downsample points \cite{sun2019novel,sun2021novel}. The non-learning clustering algorithm is also proved useful for some standard tasks like semantic segmentation \cite{kong2019pass3d} and object detection \cite{zhao2020fusion}.

In this paper, we propose to only rely on an existing neural network for the semantic classification part, then process the point-wise clustering part with traditional LiDAR cluster algorithms. As far as we know, we are the first to propose solving panoptic segmentation with the traditional point cloud clustering algorithm. Therefore, instead of developing new techniques, we would rather conduct a comprehensive technical review to investigate the performance of all existing methods. Specifically, we pick the state-of-the-art semantic model Cylinder3D \cite{zhou2020cylinder3d,zhu2021cylindrical} to provide the semantic label of each point. Then we run various LiDAR clustering algorithms to obtain the instance label of each object. Cylinder3D has open-source code with pre-trained checkpoints, thus we can easily use the same semantic model to conduct a fair comparison among different clustering methods. 

Under this setting, we evaluate various representative cluster methods implemented by us on the well-known semanticKITTI dataset \cite{behley2019semantickitti, milioto2020lidar}. We discuss more details about each selected clustering method in Section III. The same as the standard panoptic segmentation, $PQ$ (panoptic quality) is used to measure the panoptic performance of the clustering method. We also report other indicators used in SemanticKITTI when compare with methods on the leaderboard. 

\bfsection{Contributions} We believe this survey paper will be beneficial for both academic research and industrial applications. We summarize the contributions of this paper below: 

\begin{itemize}
\item \emph{We propose a new framework for LiDAR panoptic segmentation.} We are the first to demonstrate the possibility of solving the LiDAR panoptic segmentation with a semantic network and a traditional clustering method. This solution outperforms all end-to-end network solutions published recently. The classic clustering method runs in millisecond-level on CPU, alleviates labeling effort on the instance part, and has a chance to adapt to new unseen scenarios better because it has no bias toward the training set. As a traditional method, it is not a network-style black-box, thus gives people a clear idea of when the method performs better or worse.

\item \emph{We set up a benchmark for LiDAR clustering algorithms.} An issue of previous research about the LiDAR clustering method is the ambiguity of evaluation metrics due to the existence of large non-object surfaces such as walls and ground. In this paper, we filter out all those non-object points by using a semantic model with published code and checkpoints \footnote{https://github.com/xinge008/Cylinder3D}. So the clustering algorithm can only focus on points that belong to objects. We further use the well-recognized evaluation metrics \cite{song2017semantic,porzi2019seamless} of panoptic segmentation to directly evaluate and compare the effectiveness of different clustering methods. Thanks to the high quality of the SemanticKITTI dataset and leaderboard \cite{behley2019semantickitti, milioto2020lidar}, future research about LiDAR clustering topic can follow our benchmark as the de facto comparison baseline.   
\end{itemize}

\section{Literature Review}
In this section, we give a brief literature review of the panoptic segmentation task as well as the point cloud cluster method. More technical details about selected cluster methods will be introduced in the next section.

\subsection{Panoptic Segmentation}
Panoptic segmentation is a newly proposed task to fully understand a single image frame \cite{kirillov2019panoptic}. A solution to panoptic segmentation is expected to simultaneously classify pixels that belong to stuff as well as recognize each individual countable thing. The uncountable amorphous region of identical texture is known as stuff, for example, road, water, sky, etc. Things are those countable objects like persons, cars, etc. As this task is a combination of semantic segmentation and instance segmentation, it is natural to extend many deep learning models originated from existing semantic or instance solutions with extra modifications to meet the requirement of the panoptic benchmark \cite{cheng2020panoptic, wang2020axial}.

\subsection{LiDAR Panoptic Segmentation}

LiDAR panoptic segmentation is a counterpart of image panoptic segmentation on the point cloud. The model needs to give the semantic label for each point and group points that belong to the same instance together \cite{milioto2020lidar}. The range image representation gives the convenience of directly modifying image-based methods on the point cloud, including both the one-stage DeepLab style method \cite{gasperini2021panoster} and two-stage Mask R-CNN style method \cite{sirohi2021efficientlps}. Considering specific 3D information encoded in point cloud inspired some specially designed network structures, such as a dynamic shifting module developed in \cite{hong2021lidar}. 

As a very new task \cite{behley2019semantickitti,milioto2020lidar} proposed in this deep learning era, many researches directly dive into the deep learning solutions. However, despite the semantic classification part, point cloud clustering is a long-existing research topic that also has a chance to contribute as part of the panoptic task.

\subsection{Point Cloud Cluster Methods}

The point cloud is a common representation of the 3D world. How to cluster objects from the point cloud is a long-standing problem in the literature. Instead of the hot topic of dealing with 3D points using neural networks, we investigate traditional geometry-based point cloud clustering methods. We want to show those underestimated classic methods are valuable assets for solving real-world computer vision challenges if they are properly combined with recent deep learning models.

\bfsection{Clustering with Euclidean Distance} Using the Euclidean distance to cluster points is a straightforward idea explored in \cite{klasing2008clustering}, authors developed a radially bounded nearest neighbor (RBNN) algorithm for the general point cloud. They further extended RBNN by considering the normal vector \cite{klasing2009realtime}, which makes the algorithm prone to segment surfaces. Since LiDAR is mostly used in the outdoor scenario, segmenting the road surface is crucial. A novel ground segmenting algorithm was proposed in \cite{douillard2011segmentation}, other non-ground points were clustered with voxelized Euclidean neighbors. In \cite{held2016probabilistic}, researchers provided a probabilistic framework to incorporate not only the Euclidean spatial information but also the temporal information from consecutive frames. The under-segmentation error and over-segmentation error used in \cite{held2016probabilistic} also shed a light on how we should compare and evaluate various point cloud clustering methods.  

\bfsection{Clustering with Supervoxels or Superpoints} Without human knowledge, it is hard to define objects from the raw point cloud. Inspired by the concept of superpixels from the traditional image processing \cite{achanta2012slic}, some researchers are interested in finding super voxels in the Euclidean space. In \cite{papon2013voxel}, authors proposed a voxel cloud connectivity segmentation (VCCS) method which extends the definition of distance used in the iterative image pixel cluster SLIC \cite{achanta2012slic}. From the view of clustering, super voxels or super points usually over segment objects as presented and discussed in \cite{ben2018graph,landrieu2019point}.

\bfsection{Clustering on Range Image} Besides the naive Euclidean distance \cite{klasing2008clustering,klasing2009realtime,douillard2011segmentation}, researchers explored more clues aiming at finding better criteria to separate neighbor points belong to different clusters. In \cite{bogoslavskyi2016fast, yuan2019unsupervised}, the angle formed by two adjacent laser beams is considered to construct the discriminator. To make the algorithm fast enough for real-time applications, authors of \cite{bogoslavskyi2016fast} worked on the 2D range image representation of the LiDAR point cloud. This led to several works which proposed fast clustering methods on LiDAR range images by borrowing ideas from connected-component labeling (CCL) algorithms\cite{zermas2017fast,hasecke2021flic}. The connected-component labeling (CCL) is a graph algorithm used in computer vision to detect connected regions in binary digital images \cite{he2017connected}. Due to the inherent difference between binary images and LiDAR range images, authors of \cite{zermas2017fast,hasecke2021flic} modified existing CCL methods, such as two-pass CCL \cite{wu2009optimizing} or run-based CCL \cite{he2008run}, so that the clustering algorithms can run at millisecond level.      
 
\bfsection{Post-Process for Over-Segmentation} Over-segmentation means a single object is wrongly clustered into multiple parts. Successfully merging over segmented parts will improve the clustering quality for some models. In \cite{shin2017real}, a Gaussian process regression method is proposed to improve the accuracy as a post-process step. Some heuristic methods inspired by the 3D information are also attempted in \cite{li2020insclustering,li2020coarse}. Those heuristic methods mainly consider the fact that two objects can not stay too close.

\bfsection{Evaluation Metrics} Earlier publications in this field only evaluated the clustering method on private datasets with unique settings \cite{klasing2008clustering,klasing2009realtime,douillard2011segmentation}. Recent papers \cite{held2016probabilistic,shin2017real,zermas2017fast,yuan2019unsupervised,hasecke2021flic} tried to use the popular KITTI dataset \cite{geiger2013vision}. However, there are still some problems in those papers. For example, \cite{held2016probabilistic} evaluated cluster performance on the KITTI tracking dataset but only considered objects within 15 meters.

The lack of a well-recognized benchmark is a problem that limits further research about point cloud clusters. In this paper, we propose to utilize the panoptic segmentation task on SemanticKITTI as a benchmark. All the clustering methods should share the same semantic segmentation model to let the cluster only focus on points that belong to instances. We use the open-sourced state-of-the-art Cylinder3D \cite{zhou2020cylinder3d,zhu2021cylindrical} as the semantic model in this paper. A series of indicators designed for panoptic segmentation, including $PQ$, $PQ^{\dagger}$, etc, are able to measure and compare different cluster solutions as they share the same semantic part.

\section{Selected Reviewed Methods}
Existing point cloud clustering methods can be broadly summarized as four types, the Euclidean-based cluster in 3D space, clustering point cloud into supervoxel or superpoint, modified one-pass connected-component labeling on range image and modified two-pass connected-component labeling on range image. In this method review section, we pick the most typical algorithm in each type and give a more detailed introduction.

\subsection{Euclidean Cluster}
Euclidean cluster is a straightforward clustering method. It firstly constructs the kd-tree on the entire point cloud, then groups all neighbor points within a radius threshold as one instance. We illustrate the Euclidean cluster in Alg. \ref{alg:EC}. There is one parameter, the radius threshold. Intuitively, a larger threshold will group close objects together, but a smaller threshold is more sensitive to object parts with few points. Thus we choose a moderate 0.5 meter as the threshold in this paper. More details about the Euclidean cluster can be referred in \cite{rusu2010semantic,rusu20113d}. 

\begin{algorithm}[H]\footnotesize
	\SetAlgoLined
	\SetKwInOut{Input}{Input}\SetKwInOut{Output}{Output}
    \Input{Point cloud $\mathbf{P}$, Euclidean distance threshold $\mathbf{d_{th}}$}
	\Output{A list of labels for each point $\mathbf{C}$}
	Create a kd-tree to represent $\mathbf{P}$;
	\\
	$\mathbf{C}=list([0,..,0])$, $label=1$;
	\\
    \For{point $P_{i}$ in $\mathbf{P}$}{
	    \If{$C_{i}>0$}{Continue;}
	    $\mathbf{S(P_{i})} \leftarrow$ a set of points near $p_{i}$ in a sphere with radius $r < d_{th}$;
	    \\
	        $C_{i}=label$;
	        \\
	        \For{point $P_{i}$ in $\mathbf{S(P_{i})}$}{
    	        \If{$C_{i}==0$}{
    	         $C_{i}=label$;
    	        }
	        }
	        $label=label+1$;
	        }
	 \Return{$\textbf{C}$}
	\caption{ Euclidean Cluster}\label{alg:EC}
\end{algorithm}

\subsection{Supervoxel Cluster}
Superpixel SLIC \cite{achanta2012slic} is a well-known traditional image processing operator that groups local pixels to a larger one with common characteristics. Supervoxel \cite{papon2013voxel} is designed on RGB-D point cloud as a counterpart of superpixel on the 2D image. 

Compared with superpixel, there are three major differences of supervoxel. The first one is about initial seeds. In supervoxel, the seeding of clusters is done by partitioning 3D space, rather than the projected image plane. The second difference is an extra constrain that the iterative clustering algorithm enforces strict spatial connectivity of occupied voxels when considering points for clusters. The third one is the definition of the distance used in k-means. In supervoxel, instead of the distance on 2D image, the 3D Euclidean distance and the angle of the normal vector are further considered along with the color similarity. Note that the definition of supervoxel distance in point cloud library (PCL) \cite{rusu20113d} is different with original paper \cite{papon2013voxel}. We choose the one implemented in PCL in this paper.

The distance $D$ is defined in Eq. \ref{eqn:distance}. The spatial distance $D_s$ is normalized by the seeding resolution, color distance $D_c$ is the euclidean distance in normalized RGB space, and normal distance $D_n$ measures the angle between surface normal vectors. $w_c$, $w_s$, and $w_n$, are the color, spatial, and normal weights, respectively.  
\begin{align}\label{eqn:distance}
D=\sqrt{w_{c}D_{c}^{2}+\frac{w_{s}D_{s}^{2}}{3R^{2}}+w_{n}D_{n}^{2}}.
\end{align}

In this paper, we are working on the LiDAR point cloud without RGB color information, thus the color distance $D_{c}$ is set as zero for all points. The left iterative k-means of supervoxel is the same as superpixel SLIC \cite{achanta2012slic} on 2D images. Eq. \ref{eqn:distance} helps the method balance the local normal and the local Euclidean distance.

\subsection{Depth Cluster}

\setlength{\columnsep}{10pt}%
\begin{wrapfigure}{r}{.5\columnwidth}
 \includegraphics[width=\linewidth,height=40mm]{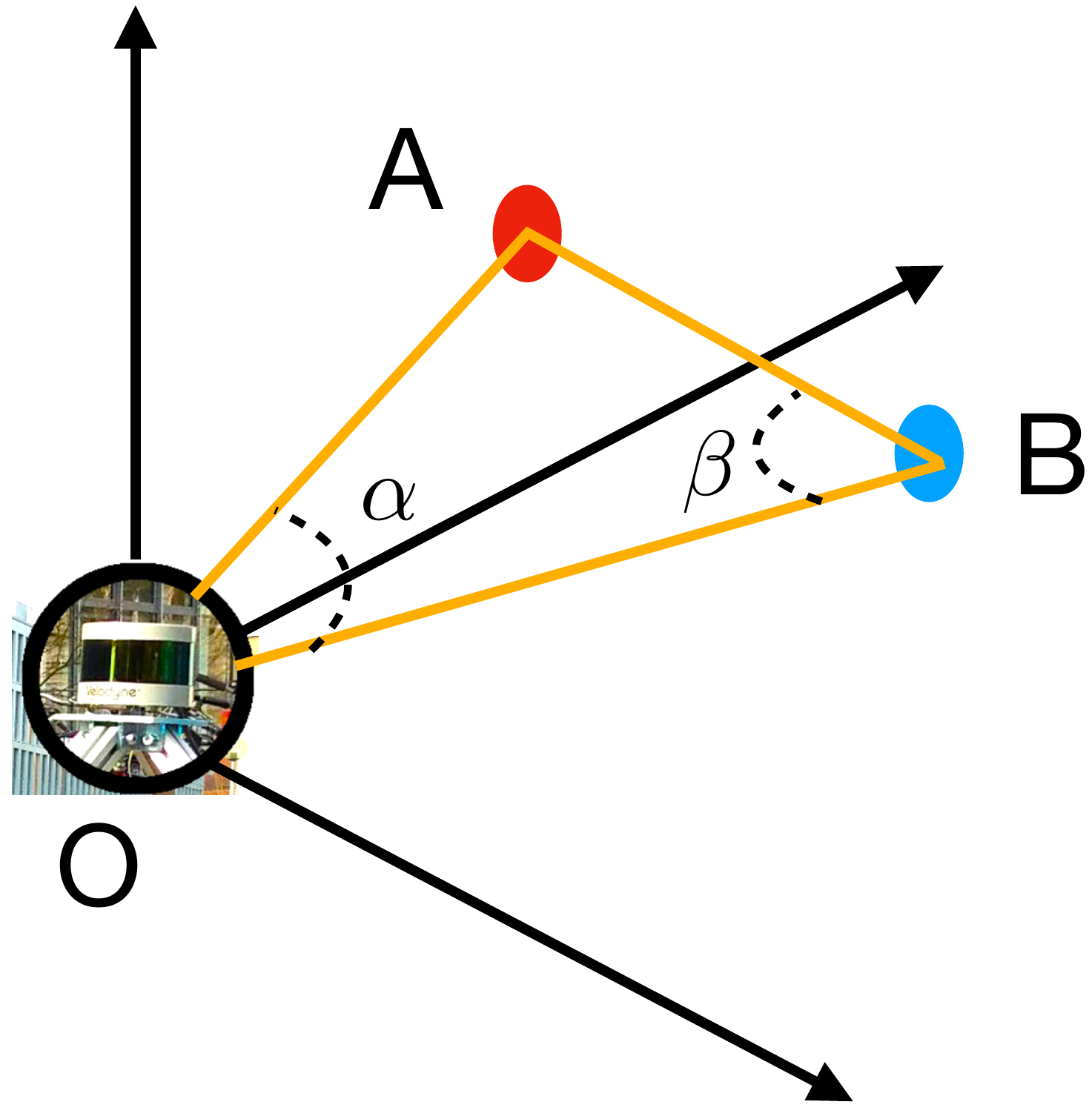}
 \caption{\small For two neighbor points A and B emitted from LiDAR O, whether the angle $\beta$ between line $BA$ and line $BO$ is larger or smaller than a threshold $\theta$ is the condition to determine if point $A$ and $B$ coming from the same object or not.}
        \label{fig:depth_cluster_condition}
\end{wrapfigure} 

Depth cluster \cite{bogoslavskyi2016fast} is a fast one-pass CCL (connected-component labeling) algorithm on the LiDAR range image. The CCL algorithm on the binary image needs to check if two neighbor pixels both have intensity one. However, the CCL on the LiDAR range image is required to define the condition that determines if two neighbor points are coming from the same object. In depth cluster algorithm, this condition is defined by using a magical angle $\beta$ shown in Fig. \ref{fig:depth_cluster_condition}. Authors \cite{bogoslavskyi2016fast} argue if the $\beta$ is larger than an angle threshold $\theta$ then point $A$ and point $B$ are coming from the same object. With this condition, the depth cluster algorithm is summarized in Alg. \ref{alg:depth_cluster}. In \cite{bogoslavskyi2016fast}, they choose $\theta=10^{o}$ as the threshold value that is also used as a fixed parameter in this survey.

\begin{algorithm}[H]\footnotesize
	\caption{ Depth Cluster}\label{alg:depth_cluster}
	\SetAlgoLined
	\SetKwInOut{Input}{Input}\SetKwInOut{Output}{Output}
    \Input{Range image $\mathbf{R}$}
	\Output{Label image $\mathbf{L}$}
	\BlankLine
	
	$\mathbf{L} \leftarrow$ $zeros(R_{rows}\times R_{cols})$,\hspace{0.3cm} $label\leftarrow 1$;\\
    \For{$r=1,...,R_{rows}$}{
     \For{$c=1,...,R_{cols}$}{
     \If{$L(r,c)==0$}{
      $\textbf{LabelComponentBFS}(r,c,label)$;\\
      $label=label + 1$;\\
      }
     }
     }
     \Return{$\mathbf{L}$}

    \setcounter{AlgoLine}{0}
    \SetKwProg{myproc}{Procedure}{\hspace{0.1cm}$\textbf{LabelComponentBFS}(r,c,label)$}{end}
    \myproc{}{
        $queue.push(\{r,c\})$;\\
        \While{$queue$ is not empty}{
        
        $\{r,c\}\leftarrow queue.top()$;\\
        $L(r,c) \leftarrow label$;\\
        \For {$\{r_n,c_n\} \in Neighborhood\{r,c\}$}{
        $d_1 \leftarrow max(R(r,c),R(r_n,c_n))$;\\
        $d_2 \leftarrow min(R(r,c),R(r_n,c_n))$;\\
        \If{$arctan \frac{d_2 sin \alpha}{d_1-d_2 cos \alpha}>\theta $}{
        $queue.push(\{r_n,c_n\});$
        }
        }
         $queue.pop()$
         }
     }

\end{algorithm}

%

\subsection{Scan-line Run Cluster}

\begin{figure}
    \begin{center}
    \includegraphics[width=\linewidth, height=5.5cm]{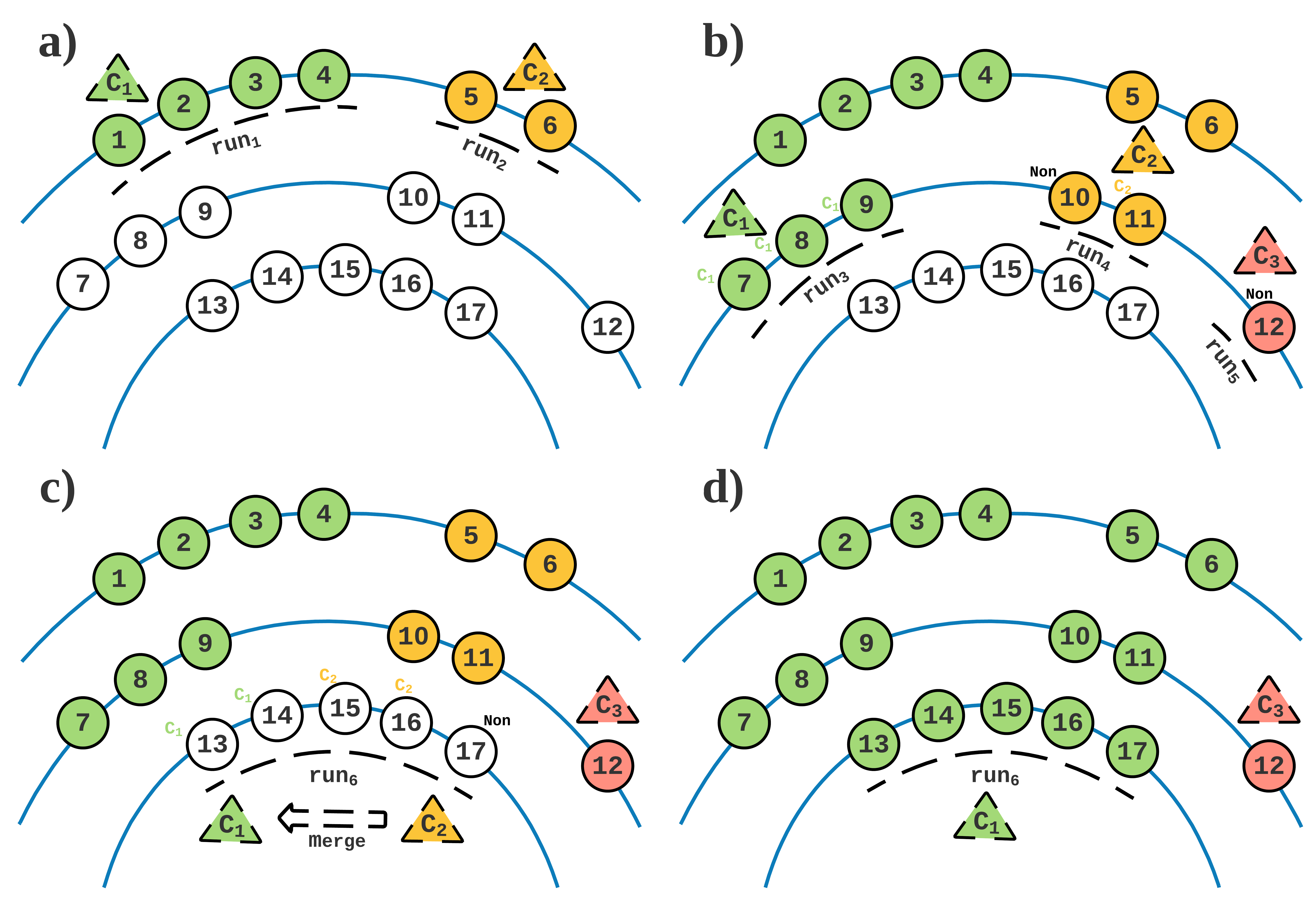}
    \caption{This figure illustrates how Scan-line Run cluster the point cloud. \textbf{a)} The first scan-line is segmented as two runs, denoted as $run_1$ and $run_2$ with labels $C_{1}$ and $C_{2}$ respectively. \textbf{b)} The second scan-line is firstly segmented as $run_3$, $run_4$ and $run_5$, then merged with existing clusters labeled as $C_{1}$ and $C_{2}$ from the previous line. The $run_5$ does not fit merging condition with previous clusters, thus receives a new cluster label $C_{3}$. \textbf{c)} When two clusters meet in the third line, they will be merged with the smaller cluster label. \textbf{d)} The result after merging two clusters.}
    \label{fig:SLR_Cluster}
    \vspace{-0.4cm}
    \end{center}
\end{figure}

Scan-line Run (SLR) cluster \cite{zermas2017fast} is a row-wise fast scan algorithm based on organized point cloud or range image. This method is a counterpart of the image-based two-pass connected component labeling (CCL) algorithm on the LiDAR range image. In SLR, all points emitted from the same horizontal angle are recognized as one scan-line. In a single scan-line, all nearby points closer than a threshold $Th_{run}$ are grouped together, called a run.

In the beginning, SLR takes in the first line, then groups all nearby points with the Euclidean distance smaller than a threshold $Th_{run}$ together as a run. Each run is assigned a unique label as initial clusters. Next, the SLR moves to the second line repeat the run segmenting and check if the new run in the second line meets the merging condition defined with a new threshold $Th_{merge}$. Two runs will be merged together if they satisfy the merging condition. The label will also propagate. If a new run in the second line does not fit the merging condition with any previous runs, a new cluster label will be assigned. For the case, if two clusters meet in a new line, SLR will merge them with the smaller cluster id. This procedure will keep moving line by line till all LiDAR scan lines are processed. We visualize the procedure in Fig. \ref{fig:SLR_Cluster} by considering the first three lines.

The algorithm is summarized in Alg. \ref{alg:Scan-line Run Cluster}. The FindNearestNeighbor function is aiming at searching the closest point in the previous scan-line. The original paper \cite{zermas2017fast} provided several ways to query the nearest neighbor with pros and cons. More details about the SLR cluster can be referred to it \cite{zermas2017fast}.

\begin{algorithm}[H]\footnotesize
	\SetAlgoLined
	\SetKwInOut{Input}{Input}\SetKwInOut{Output}{Output}
    \Input{Organized $N$ line Point Cloud $\mathbf{P}=\{P_{1},P_{2},...,P_{N}\}$ }
	\Output{Cluster ID of each point $\mathbf{L}=\{L_{1},L_{2},...,L_{N}\}$}
	
	\textbf{Initilization} \\
	$P_{i}$: $i\-th$ ordered scan-line with range value of each point\\
	$L_{i}$: cluster id of each scan-line, initialized with zero\\
    $th_{run}\leftarrow 0.5m$ : threshold to group points to a run \\
    $th_{merge}\leftarrow 1.0m$ : threshold to merge runs to a cluster \\
    $label_{idx}$ : vector of point index, each label has an independent row \\
    \BlankLine
    
    $runsCurrent = \mathbf{FindRuns}(P_{1},th_{run})$; \\
    \For{$run$ $\mathbf{in} \ $runsCurrent}
            {
                $label_{idx}.push\_back((i,run))$ ; \\
                $L_{1}[run] = label_{idx}.size$;
            }
    $runsAbove = runsCurrent$ ;\\
    \For{$i = 2: N$}{
        $runsCurrent = \mathbf{FindRuns}(P_{i}, th_{run})$; \\
         $\mathbf{UpdateLabelIndex}(runsCurrent,runsAbove)$;\\
         $runsAbove = runsCurrent$  ;
    }
    \hspace{0.2cm}Assign label from $label_{idx}$ to L;\\
    \Return{$\mathbf{L}$}
        \\
        \\
    \setcounter{AlgoLine}{0}
    \SetKwProg{myproc}{Procedure}{\hspace{0.1cm} $\textbf{FindRuns}(P_{i},\hspace{0.1cm} th_{run})$}{Return\hspace{0.1cm}$run\_all$}
    \myproc{}{
    $run\_all \leftarrow [\hspace{0.1cm}], \hspace{0.1cm} run\_each \leftarrow [\hspace{0.1cm}];$\\
     \For{$index$ $\mathbf{in}$ $P_{i}.size$}{
        \If{ $is\_empty(run\_each)$}{
            $run\_each.append(index)$;  \hspace{0.1cm} $continue$;
        }
        \If{$abs(P_{i}[index]-P_{i}[index-1])<th_{run}$}{
        $run\_each.append(index)$;\hspace{0.2cm}$\textbf{else}$\\
        $run\_all.append(run\_each)$; $run\_each\leftarrow [\hspace{0.1cm}]$;}
        }
     }

   \setcounter{AlgoLine}{0}
    \SetKwProg{myproc}{Procedure}{\hspace{0.1cm} $\textbf{UpdateLabelIndex}(runsCurrent, runsAbove)$}{end}
    \myproc{}{
     \For{$run$ $\mathbf{in} \ currentRuns$}{
        \For{$index$ $\mathbf{in} \ run$}{
            $index_{nn}= \mathbf{FindNearestNeighbor}(index,runsAbove)$; \\
            \If{$abs(P_{i}[index]-P_{i-1}[index_{nn}])<th_{merge}$}{
                $L_{i}[index]= L_{i-1}[index_{nn}]$;
            }
        }
        \If{$sum(L_{i}[run])==0$}{
             $label_{idx}.push\_back((i,run))$; \\
             $L_{i}[run]=label_{idx}.size$;
        }

        $runLabel = min(L_{i}[run])$;\\
        $L_{i}[run]=runLabel$;\\
        \For {$eachLabel$ $\mathbf{in}$ $L_{i}[run]$}{
        \If{$eachLabel>runLabel$}{
        $mergedLabel=label_{idx}[eachLabel{-}1]$;\\
        $label_{idx}[eachLabel{-}1].delete$;\\
        $label_{idx}[runLabel{-}1].insert(mergedLabel)$;\\
          }
        }
    }
     }

	\caption{Scan-line Run}\label{alg:Scan-line Run Cluster}
\end{algorithm}

\subsection{Implementations}
We implement all aforementioned four methods with C++ and wrap implementations with Pybind11 \cite{jakob2017pybind11} as python functions. Although those methods are designed for point cloud clustering, none specifically focuses on panoptic segmentation. To make them adapt better to the SemanticKITTI dataset, some minor modifications are introduced.

\bfsection{Euclidean Cluster} We directly use functions in PCL (point cloud library)\cite{rusu20113d} to implement the Euclidean cluster. Due to the use of the kd-tree data structure, the time complexity of the Euclidean cluster is non-linearly depending on the total number of points. We subsample the whole point cloud, and randomly choose one point from each $10cm\times10cm\times10cm$ voxel. The same instance label will be assigned to other points from the same voxel.

\bfsection{Supervoxel Cluster} We discard the RGB color distance as we are working on the LiDAR point cloud. 

\bfsection{Depth Cluster} The depth cluster assumes each point on the range image will have neighbors on adjacent pixels. However, the original data provided by KITTI is an unordered 3D point cloud. The range image generated by mapping this point cloud will contain many holes. To overcome this problem, our implementation will keep searching the neighbor within a threshold until finding the neighbor point.

\bfsection{Scan-line Run Cluster} Scan-line Run cluster will meet the same problem as Depth Cluster that many holes exist between LiDAR points. Therefore, we develop a make-up search strategy. The make-up search will brute-force scan every point from the above two lines, instead of the original above one line, to find the nearest points. The make-up search is triggered when searching in above one line failed, thus only slightly increases the time cost.    

\section{Experiments}

In this section, we firstly compare all four methods reviewed in this paper on the validation set of the panoptic segmentation challenge. Then, we report the performance of the best method on the test leaderboard and demonstrate a state-of-the-art result by comparing it with other end-to-end neural network solutions.

\subsection{Dataset}

Datasets used in previous clustering papers have some common problems. For example, the definition of the cluster is ambiguous, trees and walls are also clustered \cite{bogoslavskyi2016fast}; the metric is limited such as only consider objects within 15 meters \cite{held2016probabilistic}. In this paper, we propose to use the newly released panoptic segmentation challenge on SemanticKITTI as a benchmark. This dataset overcomes all previous problems. The definition of the object is very clear. If the same semantic segmentation model is used to provide labels for those non-object points, metrics designed for panoptic segmentation can exactly be used to evaluate clustering performance.

SemanticKITTI \cite{behley2019semantickitti} is a point cloud dataset about outdoor autonomous driving. It provides the benchmark of three tasks, point cloud semantic segmentation, point cloud panoptic segmentation \cite{behley2020benchmark}, and semantic scene completion \cite{behley2021towards}. The dataset contains a training split with ten LiDAR sequences, a validation split with one sequence, and a test split with eleven sequences. All labels on the test split are unavailable. Users need to submit the prediction results to the leaderboard for final evaluation scores.

\bfsection{Semantic Predictions} The clustering method works as a post-process step only focusing on instance points. All methods should use the same semantic model to make a fair comparison. Intuitively, a semantic model with better performance should give a better performance in terms of the panoptic. Considering this, we choose Cylinder3D \cite{zhou2020cylinder3d,zhu2021cylindrical} as the semantic model to provide semantic segmentation predictions. Cylinder3D is the state-of-the-art semantic model with open-sourced code. In the future, if more powerful semantic models are developed, all panoptic results reported in this paper may become even better.

\subsection{Metrics}

The metric used in this paper is the same as panoptic segmentation. $PQ$ (panoptic quality) is the major indicator defined to measure both the semantic segmentation for stuff and the instance segmentation for things \cite{kirillov2019panoptic}. An improved $PQ^{\dagger}$ is further defined \cite{porzi2019seamless}. More indicators are reported in the leaderboard, please refer to those papers for more details \cite{kirillov2019panoptic,porzi2019seamless}. We give some discussions while comparing with others in Section 4.4. Note, sharing the same semantic segmentation makes it possible to measure cluster algorithms by using those panoptic segmentation indicators. Better metrics to measure the cluster performance may be designed in the future.

\subsection{Cluster Performance Comparison}
\begin{table}[ht]
\begin{center}
    \setlength{\tabcolsep}{5.pt}
          \renewcommand{\arraystretch}{1.2} 
\begin{tabular}{c|c|c}
 \hline
   Methods & Settings &$PQ$ \\
   \hline
   \hline
    Euclidean Cluster & $d_{th}$=0.5m & 56.9\\

    \hline
    Supervoxel Cluster & $w_{c},w_{s},w_{n}$= 0.0, 1.0, 0.0 & 52.8\\
     Supervoxel Cluster & $w_{c},w_{s},w_{n}$= 0.0, 1.0, 0.5 & 52.7\\
        \hline
    Depth Cluster & $\theta$= $10^{o}$ & 55.2\\
            \hline
   Scan-line Run & $th_{run},th_{merge}$= 0.5, 1.0& \textbf{57.2}\\
   \hline
\end{tabular}
\vspace{-0.8em}
\end{center}
\caption{Performance comparison on validation set (sequence 08) of SemanticKITTI panoptic segmentation.}
\label{tab:3}
  \vspace{-0.3em}
\end{table}

\begin{table*}
\begin{center}
    \caption{The performance comparison on SemanticKITTI test set (sequence 11 to sequence 21). The red indicates the best, and the blue indicates the runner-up. sem+box means semantic segmentation with 3D object detection. sem+cluster means semantic segmentation with clustering algorithm. }
        \label{tab:table1}
    \setlength{\tabcolsep}{1.pt}
      \renewcommand{\arraystretch}{1.5} 
 \begin{tabular}{c|c|c|c c c c| c c c| c c c |c } 
  \hline
 Methods & Type & Venue& $\mathbf{PQ}$\hspace{0.1cm} & $\mathbf{PQ^{\dagger}}$ \hspace{0.1cm}& $RQ$\hspace{0.1cm} & $SQ$\hspace{0.1cm} &$PQ^{Th}$ & $RQ^{Th}$ & $SQ^{Th}$ & $PQ^{St}$ & $RQ^{St}$ &  $SQ^{St}$& $mIoU$\\ 
 \hline
 \hline
KPConv\cite{thomas2019kpconv}+PointPillars\cite{lang2019pointpillars}&sem+box &-& 44.5& 52.5& 54.4& 80.0 & 32.7& 38.7 &81.5 & 53.1 &65.9 &79.0&58.8\\
 
RangeNet++\cite{milioto2019rangenet++}+PointPillars\cite{lang2019pointpillars}&sem+box &-& 37.1& 45.9& 47.0& 75.9 & 20.2& 25.2 &75.2 & 49.3 &62.8 &76.5&52.4\\

KPConv\cite{thomas2019kpconv}+PV-RCNN \cite{shi2020pv}&sem+box &-& 50.2& 57.5& 61.4& 80.0 & 43.2& 51.4 &80.2 & 55.9 &68.7 &\textcolor{blue}{79.9}&\textcolor{blue}{62.8}\\

Panoptic RangeNet\cite{milioto2020lidar}&panoptic &iros20& 38.0& 47.0& 48.2& 76.5 & 25.6& 31.8 &76.8 & 47.1 &60.1 &76.2&50.9\\

Panoster\cite{gasperini2021panoster}&panoptic &ral21& 52.7& 59.9& 64.1& 80.7 & 49.4& 58.5 &83.3 & 55.1 &68.2 &78.8&59.9\\

PolarNet\_seg\cite{zhou2021panoptic}&panoptic &cvpr21& 54.1& 60.7& 66.0& 81.4 & \textcolor{blue}{53.3}& 60.6 &\textcolor{red}{87.2} & 54.8 &68.1 &77.2&59.5\\

DS-Net\cite{hong2021lidar}&panoptic &cvpr21& \textcolor{blue}{55.9}& \textcolor{blue}{62.5}& \textcolor{blue}{66.7}& \textcolor{red}{82.3} & \textcolor{red}{55.1}& \textcolor{red}{62.8} &\textcolor{red}{87.2} & \textcolor{blue}{56.5} &\textcolor{blue}{69.5} &78.7&61.6\\
\hline
Cylinder3D\cite{zhu2021cylindrical,zhou2020cylinder3d}+SLR \cite{zermas2017fast}&sem+cluster &-& \textcolor{red}{56.0}& \textcolor{red}{62.6}& \textcolor{red}{67.4}& \textcolor{blue}{82.1} & 51.8& \textcolor{blue}{61.0} &\textcolor{blue}{84.2} & \textcolor{red}{59.1} &\textcolor{red}{72.1} &\textcolor{red}{80.6}&\textcolor{red}{67.9}\\
 \hline
\end{tabular}
\end{center}
    \vspace{-4mm}
\end{table*}

\begin{figure*}
    \begin{center}
    \includegraphics[width=\linewidth, height=10cm]{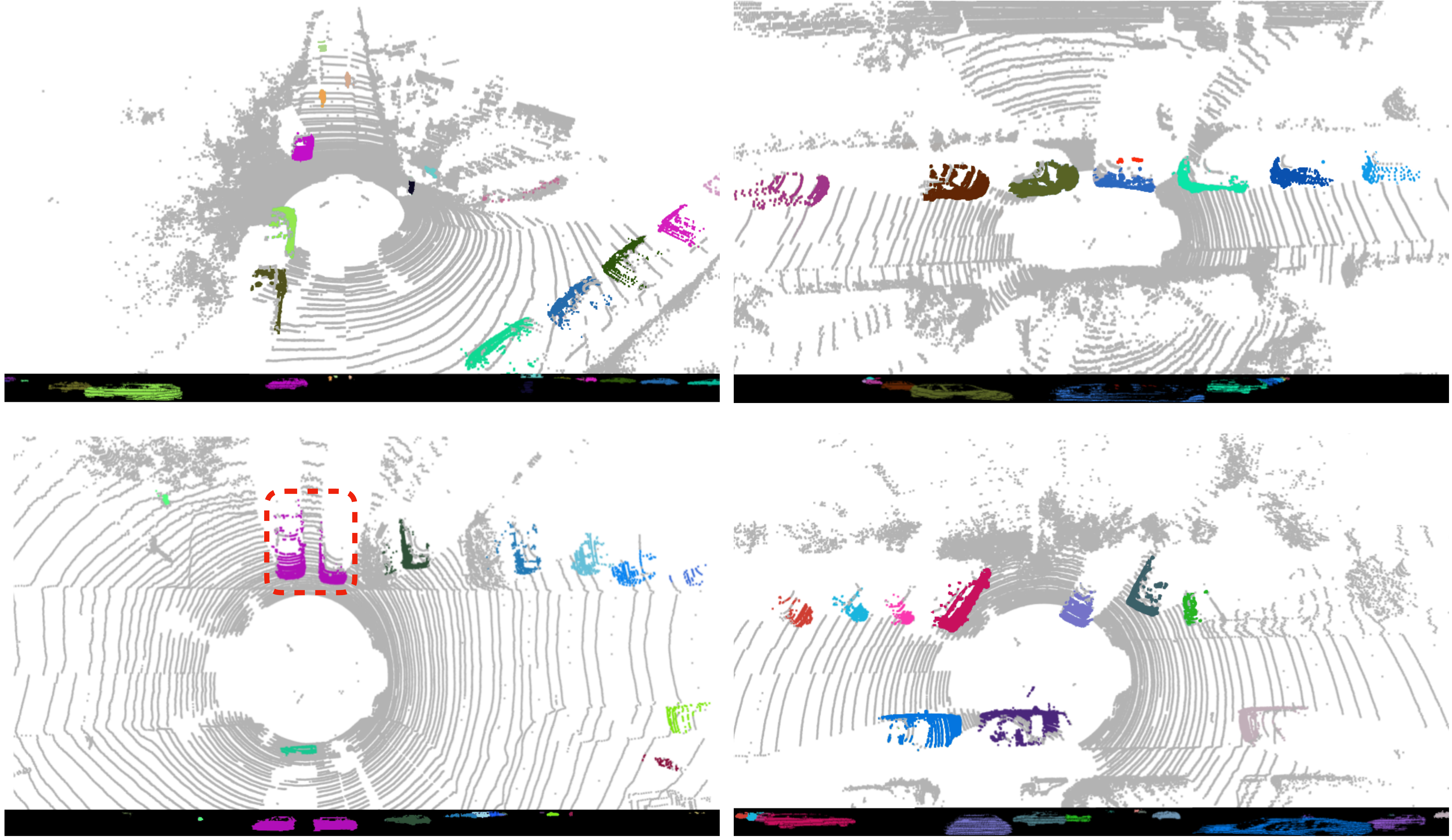}
    \caption{Visualization of the Scan-line Run clustering result on some samples from the SemanticKITTI validation set. One failure mode is two cars stay close to each other as well as park perpendicular to the LiDAR scanner, circled with the red dash box. For each sample pair, the top is the point cloud and the bottom is the range image. }
    \label{fig:example}
    \vspace{-0.8cm}
    \end{center}
\end{figure*}

By using the same semantic result, we compare the performance of all four selected cluster algorithms on the SemanticKITTI validation set in Table \ref{tab:3}. For the Euclidean cluster, we set $d_{th}$ (distance threshold) with a moderate threshold 0.5 meter. This makes sense as a large $d_{th}$ will group close objects together, and a small $d_{th}$ will be sensitive to faraway objects with few points. For the supervoxel cluster, we mainly investigate the effect of fusing normal information with distance. However, the result in Table \ref{tab:3} reports a slightly worse $PQ$ when considering the extra normal information. Without normal, supervoxel will degenerate to evenly cut of the 3D space, thus has the worst performance among all four methods. For the depth cluster, we keep the same choice of the angle threshold $\theta$ as the original paper. For the scan-line run cluster, the values of two distance thresholds are also kept the same with the paper. The scan-line run cluster achieves the best performance on the validation set in all four methods. We visualize some samples of this clustering algorithm on SemanticKITTI validation set in Fig. \ref{fig:example}.

\subsection{State-of-the-art Comparison on Leaderboard}

From Table \ref{tab:3}, we show the Scan-line Run (SLR) \cite{zermas2017fast} is the best clustering algorithm among the four methods evaluated on the SemanticKITTI validation set. Then, we evaluate our pipeline on the SemanticKITTI test set by using SLR as the traditional cluster to combine with Cylinder3D \cite{zhu2021cylindrical,zhou2020cylinder3d}. In Table \ref{tab:table1}, we report the performance and compare this hybrid solution with others. There are two major types of competitors. One type is panoptic segmentation combined with semantic segmentation and 3D object detection, numbers are cited from \cite{hong2021lidar}. This kind of solution needs two large networks with redundant computational cost and information. The second type is the end-to-end panoptic segmentation model. The proposed hybrid solution outperforms all others on the leaderboard in terms of two major indicators $PQ$ and $PQ^{\dagger}$. For other metrics reported in the leaderboard, our pipeline achieves comparable performance with the current state-of-the-art method DS-Net \cite{hong2021lidar}. 

\bfsection{Relationship with Semantic Segmentation} One interesting finding from the Table \ref{tab:table1} is the number comparison with DS-Net \cite{hong2021lidar}. Our hybrid solution performs better for $PQ^{St}$, $RQ^{St}$ and $SQ^{St}$, but performs worse for $PQ^{Th}$, $RQ^{Th}$ and $SQ^{Th}$. The indicators $PQ^{St}$, $RQ^{St}$, and $SQ^{St}$ measure the model performance on non-object points. So we can say the reason our hybrid solution can achieve the state-of-the-art performance is because the traditional cluster algorithm almost keeps the same high performance of the semantic segmentation model as well as provides pretty good instance clustering results. This is a unique advantage of the traditional method. For those end-to-end network solutions, involving the extra instance branch beside the semantic segmentation makes the training of the network become a multi-task learning problem. It is unclear how multi-task learning will affect the network performance on semantic segmentation. In Cylinder3D \cite{zhou2020cylinder3d,zhu2021cylindrical}, authors show extending the semantic with extra instance branch decreases the mIoU from 65.9 to 63.5 with 2.4 point drop on the validation set. In our case, adding the traditional cluster only decreases the mIoU from 68.9 to 67.9 with one point drop on the test leaderboard. This advantage of the traditional method with the high-quality instance clustering performance helps our proposed pipeline outperform others for major panoptic indicators $PQ$ and $PQ^{\dagger}$.

\subsection{Time Complexity}
The inference time is also an important factor. Let's assume there are $N$ points in total, and the number of pixels on the range image is roughly the same as $N$. Due to the construction of the kd-tree, the Euclidean cluster has $Nlog(N)$ time complexity. We voxelize the space with the cube of size $10cm$. If there are $m$ cubes with points inside them, the time complexity of the Euclidean cluster is $mlog(m)$, and $m\ll N$ for 64 line LiDAR. Both the depth cluster and scan-line run cluster are working on the range image with modified connected-component labeling algorithms. Therefore, those two clustering methods have the linear $O(N)$ complexity. The inference speed of supervoxel is saved by limit the searching region, or seed resolution. In Table \ref{tab:hh}, we directly report the inference time of the supervoxel implementation in point cloud library (PCL) with 0.5m voxel resolution and 8.0m seed resolution.   

\begin{table}[ht]
\vspace{-0.8em}
\begin{center}
    \setlength{\tabcolsep}{8.pt}
          \renewcommand{\arraystretch}{1.2} 
\begin{tabular}{c|c}
 \hline
   Methods & Speed of Single Frame \\
   \hline
   \hline
    Euclidean Cluster & 16.2ms\\
    \hline
    Supervoxel Cluster & 62.5ms \\
        \hline
    Depth Cluster &18.1ms \\
            \hline
   Scan-line Run & 29.0ms \\
   \hline
\end{tabular}
\vspace{-0.8em}
\end{center}
\caption{Inference time of one LiDAR frame. All our implementations run on a single thread of i7-6700K CPU @ 4.00GHz.}
\label{tab:hh}
  \vspace{-0.5em}
\end{table}

We report the average single frame inference time of our implementations in Table \ref{tab:hh}. Although the inference time is also decided by the coding quality, we present those numbers here to emphasize that traditional cluster algorithms are fast enough to work with the 10Hz LiDAR frame rate. Note, the original depth cluster paper \cite{bogoslavskyi2016fast} reports 4.7ms to process each LiDAR frame. This time difference with our implementation is partially caused by the existence of many holes on the KITTI range image.

\section{Conclusion}
This is a technical survey paper about LiDAR point cloud cluster methods and their performance evaluation on the panoptic segmentation benchmark. We want to deliver the message that the traditional clustering algorithm is a useful asset for 3D point cloud tasks. From the view of the LiDAR panoptic segmentation, we propose a new hybrid pipeline with state-of-the-art performance. From the view of the point cloud cluster method, we provide a benchmark to fairly evaluate and compare different clustering algorithms. All the code used in this paper will be released to the public. We believe this work will be useful for both the academic research and engineering designs to better understand the LiDAR point cloud towards solving the real-world problem.

{\small
\bibliographystyle{ieee_fullname}
\bibliography{egbib}

\begin{thebibliography}{10}\itemsep=-1pt

\bibitem{achanta2012slic}
Radhakrishna Achanta, Appu Shaji, Kevin Smith, Aurelien Lucchi, Pascal Fua, and
  Sabine S{\"u}sstrunk.
\newblock Slic superpixels compared to state-of-the-art superpixel methods.
\newblock {\em IEEE transactions on pattern analysis and machine intelligence},
  34(11):2274--2282, 2012.

\bibitem{behley2021towards}
Jens Behley, Martin Garbade, Andres Milioto, Jan Quenzel, Sven Behnke,
  J{\"u}rgen Gall, and Cyrill Stachniss.
\newblock Towards 3d lidar-based semantic scene understanding of 3d point cloud
  sequences: The semantickitti dataset.
\newblock {\em The International Journal of Robotics Research}, page
  02783649211006735, 2021.

\bibitem{behley2019semantickitti}
Jens Behley, Martin Garbade, Andres Milioto, Jan Quenzel, Sven Behnke, Cyrill
  Stachniss, and Jurgen Gall.
\newblock Semantickitti: A dataset for semantic scene understanding of lidar
  sequences.
\newblock In {\em Proceedings of the IEEE/CVF International Conference on
  Computer Vision}, pages 9297--9307, 2019.

\bibitem{behley2020benchmark}
Jens Behley, Andres Milioto, and Cyrill Stachniss.
\newblock A benchmark for lidar-based panoptic segmentation based on kitti.
\newblock {\em arXiv preprint arXiv:2003.02371}, 2020.

\bibitem{ben2018graph}
Yizhak Ben-Shabat, Tamar Avraham, Michael Lindenbaum, and Anath Fischer.
\newblock Graph based over-segmentation methods for 3d point clouds.
\newblock {\em Computer Vision and Image Understanding}, 174:12--23, 2018.

\bibitem{bogoslavskyi2016fast}
Igor Bogoslavskyi and Cyrill Stachniss.
\newblock Fast range image-based segmentation of sparse 3d laser scans for
  online operation.
\newblock In {\em 2016 IEEE/RSJ International Conference on Intelligent Robots
  and Systems (IROS)}, pages 163--169. IEEE, 2016.

\bibitem{chen2019suma++}
Xieyuanli Chen, Andres Milioto, Emanuele Palazzolo, Philippe Giguere, Jens
  Behley, and Cyrill Stachniss.
\newblock Suma++: Efficient lidar-based semantic slam.
\newblock In {\em 2019 IEEE/RSJ International Conference on Intelligent Robots
  and Systems (IROS)}, pages 4530--4537. IEEE, 2019.

\bibitem{cheng2020panoptic}
Bowen Cheng, Maxwell~D Collins, Yukun Zhu, Ting Liu, Thomas~S Huang, Hartwig
  Adam, and Liang-Chieh Chen.
\newblock Panoptic-deeplab: A simple, strong, and fast baseline for bottom-up
  panoptic segmentation.
\newblock In {\em Proceedings of the IEEE/CVF Conference on Computer Vision and
  Pattern Recognition}, pages 12475--12485, 2020.

\bibitem{douillard2011segmentation}
Bertrand Douillard, James Underwood, Noah Kuntz, Vsevolod Vlaskine, Alastair
  Quadros, Peter Morton, and Alon Frenkel.
\newblock On the segmentation of 3d lidar point clouds.
\newblock In {\em 2011 IEEE International Conference on Robotics and
  Automation}, pages 2798--2805. IEEE, 2011.

\bibitem{gasperini2021panoster}
Stefano Gasperini, Mohammad-Ali~Nikouei Mahani, Alvaro Marcos-Ramiro, Nassir
  Navab, and Federico Tombari.
\newblock Panoster: End-to-end panoptic segmentation of lidar point clouds.
\newblock {\em IEEE Robotics and Automation Letters}, 6(2):3216--3223, 2021.

\bibitem{geiger2013vision}
Andreas Geiger, Philip Lenz, Christoph Stiller, and Raquel Urtasun.
\newblock Vision meets robotics: The kitti dataset.
\newblock {\em The International Journal of Robotics Research},
  32(11):1231--1237, 2013.

\bibitem{hasecke2021flic}
Frederik Hasecke, Lukas Hahn, and Anton Kummert.
\newblock Flic: Fast lidar image clustering.
\newblock In {\em ICPRAM}, pages 25--35, 2021.

\bibitem{he2008run}
Lifeng He, Yuyan Chao, and Kenji Suzuki.
\newblock A run-based two-scan labeling algorithm.
\newblock {\em IEEE transactions on image processing}, 17(5):749--756, 2008.

\bibitem{he2017connected}
Lifeng He, Xiwei Ren, Qihang Gao, Xiao Zhao, Bin Yao, and Yuyan Chao.
\newblock The connected-component labeling problem: A review of
  state-of-the-art algorithms.
\newblock {\em Pattern Recognition}, 70:25--43, 2017.

\bibitem{held2016probabilistic}
David Held, Devin Guillory, Brice Rebsamen, Sebastian Thrun, and Silvio
  Savarese.
\newblock A probabilistic framework for real-time 3d segmentation using
  spatial, temporal, and semantic cues.
\newblock In {\em Robotics: Science and Systems}, volume~12, 2016.

\bibitem{hendy2020fishing}
Noureldin Hendy, Cooper Sloan, Feng Tian, Pengfei Duan, Nick Charchut, Yuesong
  Xie, Chuang Wang, and James Philbin.
\newblock Fishing net: Future inference of semantic heatmaps in grids.
\newblock {\em arXiv preprint arXiv:2006.09917}, 2020.

\bibitem{hong2021lidar}
Fangzhou Hong, Hui Zhou, Xinge Zhu, Hongsheng Li, and Ziwei Liu.
\newblock Lidar-based panoptic segmentation via dynamic shifting network.
\newblock In {\em Proceedings of the IEEE/CVF Conference on Computer Vision and
  Pattern Recognition}, pages 13090--13099, 2021.

\bibitem{jakob2017pybind11}
Wenzel Jakob, Jason Rhinelander, and Dean Moldovan.
\newblock pybind11--seamless operability between c++ 11 and python.
\newblock {\em URL: https://github. com/pybind/pybind11}, 2017.

\bibitem{kirillov2019panoptic}
Alexander Kirillov, Kaiming He, Ross Girshick, Carsten Rother, and Piotr
  Doll{\'a}r.
\newblock Panoptic segmentation.
\newblock In {\em Proceedings of the IEEE/CVF Conference on Computer Vision and
  Pattern Recognition}, pages 9404--9413, 2019.

\bibitem{klasing2008clustering}
Klaas Klasing, Dirk Wollherr, and Martin Buss.
\newblock A clustering method for efficient segmentation of 3d laser data.
\newblock In {\em 2008 IEEE international conference on robotics and
  automation}, pages 4043--4048. IEEE, 2008.

\bibitem{klasing2009realtime}
Klaas Klasing, Dirk Wollherr, and Martin Buss.
\newblock Realtime segmentation of range data using continuous nearest
  neighbors.
\newblock In {\em 2009 IEEE International Conference on Robotics and
  Automation}, pages 2431--2436. IEEE, 2009.

\bibitem{kong2019pass3d}
Xin Kong, Guangyao Zhai, Baoquan Zhong, and Yong Liu.
\newblock Pass3d: Precise and accelerated semantic segmentation for 3d point
  cloud.
\newblock In {\em 2019 IEEE/RSJ International Conference on Intelligent Robots
  and Systems (IROS)}, pages 3467--3473. IEEE, 2019.

\bibitem{landrieu2019point}
Loic Landrieu and Mohamed Boussaha.
\newblock Point cloud oversegmentation with graph-structured deep metric
  learning.
\newblock In {\em Proceedings of the IEEE/CVF Conference on Computer Vision and
  Pattern Recognition}, pages 7440--7449, 2019.

\bibitem{lang2019pointpillars}
Alex~H Lang, Sourabh Vora, Holger Caesar, Lubing Zhou, Jiong Yang, and Oscar
  Beijbom.
\newblock Pointpillars: Fast encoders for object detection from point clouds.
\newblock In {\em Proceedings of the IEEE/CVF Conference on Computer Vision and
  Pattern Recognition}, pages 12697--12705, 2019.

\bibitem{li2020insclustering}
You Li, Cl{\'e}ment Le~Bihan, Txomin Pourtau, and Thomas Ristorcelli.
\newblock Insclustering: Instantly clustering lidar range measures for
  autonomous vehicle.
\newblock In {\em 2020 IEEE 23rd International Conference on Intelligent
  Transportation Systems (ITSC)}, pages 1--6. IEEE, 2020.

\bibitem{li2020coarse}
You Li, Clement Lebihan, Txomin Pourtau, Thomas Ristorcelli, and Javier
  Ibanez-Guzman.
\newblock Coarse-to-fine segmentation on lidar point clouds in spherical
  coordinate and beyond.
\newblock {\em IEEE Transactions on Vehicular Technology}, 2020.

\bibitem{milioto2020lidar}
Andres Milioto, Jens Behley, Chris McCool, and Cyrill Stachniss.
\newblock Lidar panoptic segmentation for autonomous driving.
\newblock In {\em 2020 IEEE/RSJ International Conference on Intelligent Robots
  and Systems (IROS)}, pages 8505--8512. IEEE, 2020.

\bibitem{milioto2019rangenet++}
Andres Milioto, Ignacio Vizzo, Jens Behley, and Cyrill Stachniss.
\newblock Rangenet++: Fast and accurate lidar semantic segmentation.
\newblock In {\em 2019 IEEE/RSJ International Conference on Intelligent Robots
  and Systems (IROS)}, pages 4213--4220. IEEE, 2019.

\bibitem{papon2013voxel}
Jeremie Papon, Alexey Abramov, Markus Schoeler, and Florentin Worgotter.
\newblock Voxel cloud connectivity segmentation-supervoxels for point clouds.
\newblock In {\em Proceedings of the IEEE conference on computer vision and
  pattern recognition}, pages 2027--2034, 2013.

\bibitem{porzi2019seamless}
Lorenzo Porzi, Samuel~Rota Bulo, Aleksander Colovic, and Peter Kontschieder.
\newblock Seamless scene segmentation.
\newblock In {\em Proceedings of the IEEE/CVF Conference on Computer Vision and
  Pattern Recognition}, pages 8277--8286, 2019.

\bibitem{rusu2010semantic}
Radu~Bogdan Rusu.
\newblock Semantic 3d object maps for everyday manipulation in human living
  environments.
\newblock {\em KI-K{\"u}nstliche Intelligenz}, 24(4):345--348, 2010.

\bibitem{rusu20113d}
Radu~Bogdan Rusu and Steve Cousins.
\newblock 3d is here: Point cloud library (pcl).
\newblock In {\em 2011 IEEE international conference on robotics and
  automation}, pages 1--4. IEEE, 2011.

\bibitem{shi2020pv}
Shaoshuai Shi, Chaoxu Guo, Li Jiang, Zhe Wang, Jianping Shi, Xiaogang Wang, and
  Hongsheng Li.
\newblock Pv-rcnn: Point-voxel feature set abstraction for 3d object detection.
\newblock In {\em Proceedings of the IEEE/CVF Conference on Computer Vision and
  Pattern Recognition}, pages 10529--10538, 2020.

\bibitem{shin2017real}
Myung-Ok Shin, Gyu-Min Oh, Seong-Woo Kim, and Seung-Woo Seo.
\newblock Real-time and accurate segmentation of 3-d point clouds based on
  gaussian process regression.
\newblock {\em IEEE Transactions on Intelligent Transportation Systems},
  18(12):3363--3377, 2017.

\bibitem{sirohi2021efficientlps}
Kshitij Sirohi, Rohit Mohan, Daniel B{\"u}scher, Wolfram Burgard, and Abhinav
  Valada.
\newblock Efficientlps: Efficient lidar panoptic segmentation.
\newblock {\em arXiv preprint arXiv:2102.08009}, 2021.

\bibitem{song2017semantic}
Shuran Song, Fisher Yu, Andy Zeng, Angel~X Chang, Manolis Savva, and Thomas
  Funkhouser.
\newblock Semantic scene completion from a single depth image.
\newblock In {\em Proceedings of the IEEE Conference on Computer Vision and
  Pattern Recognition}, pages 1746--1754, 2017.

\bibitem{sun2019novel}
Xuebin Sun, Han Ma, Yuxiang Sun, and Ming Liu.
\newblock A novel point cloud compression algorithm based on clustering.
\newblock {\em IEEE Robotics and Automation Letters}, 4(2):2132--2139, 2019.

\bibitem{sun2021novel}
Xuebin Sun, Yuxiang Sun, Weixun Zuo, Shing~Shin Cheng, and Ming Liu.
\newblock A novel coding scheme for large-scale point cloud sequences based on
  clustering and registration.
\newblock {\em IEEE Transactions on Automation Science and Engineering}, 2021.

\bibitem{thomas2019kpconv}
Hugues Thomas, Charles~R Qi, Jean-Emmanuel Deschaud, Beatriz Marcotegui,
  Fran{\c{c}}ois Goulette, and Leonidas~J Guibas.
\newblock Kpconv: Flexible and deformable convolution for point clouds.
\newblock In {\em Proceedings of the IEEE/CVF International Conference on
  Computer Vision}, pages 6411--6420, 2019.

\bibitem{wang2020axial}
Huiyu Wang, Yukun Zhu, Bradley Green, Hartwig Adam, Alan Yuille, and
  Liang-Chieh Chen.
\newblock Axial-deeplab: Stand-alone axial-attention for panoptic segmentation.
\newblock In {\em European Conference on Computer Vision}, pages 108--126.
  Springer, 2020.

\bibitem{wu2009optimizing}
Kesheng Wu, Ekow Otoo, and Kenji Suzuki.
\newblock Optimizing two-pass connected-component labeling algorithms.
\newblock {\em Pattern Analysis and Applications}, 12(2):117--135, 2009.

\bibitem{yuan2019unsupervised}
Xia Yuan, Yangyukun Mao, and Chunxia Zhao.
\newblock Unsupervised segmentation of urban 3d point cloud based on
  lidar-image.
\newblock In {\em 2019 IEEE International Conference on Robotics and
  Biomimetics (ROBIO)}, pages 2565--2570. IEEE, 2019.

\bibitem{zermas2017fast}
Dimitris Zermas, Izzat Izzat, and Nikolaos Papanikolopoulos.
\newblock Fast segmentation of 3d point clouds: A paradigm on lidar data for
  autonomous vehicle applications.
\newblock In {\em 2017 IEEE International Conference on Robotics and Automation
  (ICRA)}, pages 5067--5073. IEEE, 2017.

\bibitem{zhao2020fusion}
Xiangmo Zhao, Pengpeng Sun, Zhigang Xu, Haigen Min, and Hongkai Yu.
\newblock Fusion of 3d lidar and camera data for object detection in autonomous
  vehicle applications.
\newblock {\em IEEE Sensors Journal}, 20(9):4901--4913, 2020.

\bibitem{zhou2020cylinder3d}
Hui Zhou, Xinge Zhu, Xiao Song, Yuexin Ma, Zhe Wang, Hongsheng Li, and Dahua
  Lin.
\newblock Cylinder3d: An effective 3d framework for driving-scene lidar
  semantic segmentation.
\newblock {\em arXiv preprint arXiv:2008.01550}, 2020.

\bibitem{zhou2021panoptic}
Zixiang Zhou, Yang Zhang, and Hassan Foroosh.
\newblock Panoptic-polarnet: Proposal-free lidar point cloud panoptic
  segmentation.
\newblock In {\em Proceedings of the IEEE/CVF Conference on Computer Vision and
  Pattern Recognition}, pages 13194--13203, 2021.

\bibitem{zhu2021cylindrical}
Xinge Zhu, Hui Zhou, Tai Wang, Fangzhou Hong, Yuexin Ma, Wei Li, Hongsheng Li,
  and Dahua Lin.
\newblock Cylindrical and asymmetrical 3d convolution networks for lidar
  segmentation.
\newblock In {\em Proceedings of the IEEE/CVF Conference on Computer Vision and
  Pattern Recognition}, pages 9939--9948, 2021.

\end{thebibliography}
}

\end{document}